\pdfoutput=1
\documentclass[11pt]{article}

\usepackage{acl}
\usepackage{caption} 
\usepackage{xcolor}
\usepackage{times}
\usepackage{latexsym}
\usepackage{amssymb}
\usepackage{amsmath}
\usepackage{amsthm}
\usepackage{graphicx}
\usepackage{makecell}
\usepackage{subcaption}
\usepackage{tcolorbox}
\usepackage{booktabs}
\usepackage{multirow}
\usepackage[T1]{fontenc}
\usepackage[ruled,vlined]{algorithm2e}
\SetAlCapNameFnt{\scriptsize}
\SetAlCapFnt{\scriptsize}
\SetAlFnt{\scriptsize}
\usepackage{color}
\usepackage{url}

\usepackage[utf8]{inputenc}

\usepackage{microtype}

\title{GRETEL: Graph Contrastive Topic Enhanced Language Model for Long Document Extractive Summarization}

\author{Qianqian Xie \\
  National Centre for Text Mining,\\ Department of Computer Science,\\ The University of Manchester\\
  \texttt{qianqian.xie@manchester.ac.uk} \\
  \And
  Jimin Huang \\
  Chancefocus AMC. \\
  \texttt{jimin@chancefocus.com} \\
  \AND
  Tulika Saha \and Sophia Ananiadou\\
  National Centre for Text Mining, Department of Computer Science,\\ The University of Manchester\\
  \texttt{\{tulika.saha,sophia.ananiadou\}@manchester.ac.uk} \\
  }

\begin{document}
\maketitle
\begin{abstract}
Recently, neural topic models (NTMs) have been incorporated into pre-trained language models (PLMs), to capture the global semantic information for text summarization.
However, in these methods, there remain limitations in how they capture and integrate the global semantic information.
In this paper, we propose a novel model, \textbf{G}raph cont\textbf{R}astiv\textbf{E} \textbf{T}opic \textbf{E}nhanced \textbf{L}anguage model (GRETEL), that incorporates the graph contrastive topic model with the pre-trained language model, to fully leverage both the global and local contextual semantics for long document extractive summarization.
To better capture and incorporate the global semantic information into PLMs, the graph contrastive topic model integrates the hierarchical transformer encoder and the graph contrastive learning to fuse the semantic information from the global document context and the gold summary.
To this end, GRETEL encourages the model to efficiently extract salient sentences that are topically related to the gold summary, rather than redundant sentences that cover sub-optimal topics.
Experimental results\footnote{\url{https://github.com/xashely/GRETEL\_extractive}} on both general domain and biomedical datasets demonstrate that our proposed method outperforms SOTA methods.
\end{abstract}

\section{Introduction}
Due to the well-known limitation of pre-trained language models (PLMs)~\cite{devlin2019bert,wang2021pre} that they fail to capture long-range dependencies~\cite{beltagy2020longformer}, attempts have been proposed to integrate neural topic models (NTMs)~\cite{cao2015novel,peng2018neural,xie2021graph} into PLMs, which have shown significant improvement in the performance of the text summarization task~\cite{wang2020friendly, cui2021topic, nguyen2021enriching, fu2020document}.
In addition to the local contextual information captured in PLMs, NTMs can provide an approximation of the global semantics captured from document contents, i.e., latent topics, as well as their posterior topic representations.
The global semantics 
are further used to guide the model, to generate coherent summaries which cover the most relevant topics discussed within the document, via the attention mechanism~\cite{wang2020friendly, aralikatte2021focus, nguyen2021enriching, fu2020document} or graph neural networks (GNNs)~\cite{cui2021topic, cui2020enhancing}.

However, there exists the
\textbf{semantic gap} between latent topics as the approximation global semantics, and the true global semantics due to two major limitations for existing methods.
The first limitation concerns the nature of \textbf{unsupervised topic inference} in these methods, where topics and posterior topic distributions are learned from documents in an unsupervised manner, without considering the key semantic information conveyed in the gold summary (mostly the abstract). 
Existing methods using document-word features, without accessing the semantic information of the gold summary, can extract sub-optimal topics with high-frequency words.
However, sub-optimal topics with high-frequency words, do not necessarily cover the true global semantics that is condensed in the gold summary.
This results in the wrong assignment of the document with sub-optimal topics, and consequently the model extract redundant sentences containing high-frequency topic words.

Another limitation is that existing methods rely on \textbf{document word features} such as Bag-of-Words (BOWs) to extract latent topics, and disregard the sequential and syntactic dependency between words.
This may lead to sequentially and syntactically correlated words being allocated to different topics. 
One solution is to provide NTMs with contextual representations from PLMs, which is nevertheless challenging to directly apply in existing methods for text summarization. PLMs in existing methods are forced to truncate the input to a limited length owing to the complexity of language models.
Thus the representations from partial content of the document cannot necessarily help NTMs to mine informative topics that cover the whole content of the document, especially for long documents.
Overall, although existing methods encourage the summary with sentences that are topically similar to the document topic distribution, the summary focuses more on sentences with high-frequency words and can have low semantic similarity with the gold summary.
To help understand these limitations, we make a detailed analysis based on the benchmark datasets in section~\ref{sec:ana}.

In response to the above, we propose a novel \textbf{G}raph cont\textbf{R}astiv\textbf{E} \textbf{T}opic \textbf{E}nhanced \textbf{L}anguage model (GRETEL), that incorporates the graph contrastive topic model (GCTM) empowered by the semantic information of the gold summary and the global document context, with PLMs for long document extractive summarization.
The first distinguishing feature of GRETEL is the employment of the hierarchical transformer encoder (HTE) to fully embed the global context of long documents, to inform topic representations of documents and sentences.
The global contextual information captured in HTE but missing in the BOW feature, allows the model to learn more discriminative document and sentence topic representations, and coherent topics.

Secondly, it utilizes graph contrastive learning with the supervised information from the gold summary. 
It pushes close topic representations of documents and sentences that have high semantic similarity with the gold summary and pulls away otherwise.
This encourages the model to capture better global semantic information: latent topics that describe the most key information in the original document content. 
Therefore, it allows the model to select key sentences that are topically similar to the gold summary.
Experimental results demonstrate that our method can effectively distinguish salient sentences from documents with both global and local semantics, leading to superior performance compared to previous methods.


\section{Related Work}
\textbf{Topic enhanced PLMs for Text Summarization.}
Many studies have investigated the application of PLMs for extractive summarization, including BERTSum~\cite{liu2019text}, DiscoBert\cite{xu2020discourse}, MatchSum~\cite{zhong2020extractive} et al.
However, these methods fail to capture the global context of long documents due to the limitation of PLMs.
To address it, several studies combined topic modeling with PLMs to introduce global semantic information.
\citet{wang2020friendly} proposed to extract firstly latent topics independently and then use them to improve the summarization model. 
Other studies~\cite{aralikatte2021focus, nguyen2021enriching, cui2020enhancing} proposed to use the BOW as input features for neural topic modeling and improved the transformer encoder and decoder with the extracted latent topics with an attention mechanism for abstractive summarization~\cite{aralikatte2021focus, fu2020document, nguyen2021enriching}. 
\citet{cui2020enhancing, cui2021topic} proposed to use the graph neural network to infuse topics into contextual representations from PLMs, for multi-document abstractive summarization and extractive summarization.
\citet{fu2020document} considered extract both document and paragraph-level topic distribution, and use them to guide the abstractive summarization.

\textbf{PLMs for Long Document Summarization.}
To address the limitation of encoding the full context of long documents using PLMs, another direction is to design the efficient sparse self-attention or using a sliding window. 
\citet{narayan2020stepwise} proposed a step-wise model with a structured transformer.
\citet{huang2021efficient} proposed a computationally efficient method based on the head-wise positional strides, to identify salient content for long documents.
\citet{liu2021hetformer} employed a transformer with multi-granularity sparse attentions.
\citet{cui2021sliding} used a sliding selector network with dynamic memory, in which the sliding window is used to encode input documents, segment by segment.
\citet{grail2021globalizing} divided long documents into multiple blocks and encoded them by independent transformer networks. 

\section{Dataset-dependent Analysis for Limitations}
\label{sec:ana}
To better illustrate the limitations of existing methods, we present a dataset-dependent analysis, with the aim to answer two key questions:
1) Do the extracted topics from topic models tend to focus on high-frequency words? and
2) Due to it, would there be a semantic gap between topics as the approximation of global semantics and true global semantics in the gold summary?

We first present the top-10 words of topics learned by the traditional topic model LDA on the PubMed~\cite{cohan2018discourse} dataset, as shown in Table~\ref{tab:to}.
It shows that there is a high overlap between words in learned topics and high-frequency words.
This is also reported in previous studies~\cite{griffiths2002probabilistic, steyvers2007probabilistic, chi2019topic}, that words are mentioned more frequently, have a higher probability conditioned on topics on average.
Since, they infer the posterior distribution of documents over topics, according to the co-occurrences of words in the whole document collections.
\begin{table}[!hbt]
    \centering
    \scriptsize
    \resizebox{0.45\textwidth}{12mm}{
    \begin{tabular}{|p{1.0\linewidth}|}
    \hline
     T1: type treatment consistent needed lower disorders sensitive patient acid way\\
     T2: male group treatment followed cells per side plasma american health\\
     T3: al dna clinical risk observed tube lower inflammatory et features\\
     T4: type al clinical mice bacteria high vs posterior conditions side\\
     T5: differences performed results side number higher size tube et patients\\
     T6: dna revealed smoking control mental number change sd light versus\\
     \hline
     \textbf{Top high-frequency words:} patients, study, using, cells, group, treatment, et, one, al, data, studies, two, patient, results, cell, time, however, figure, significant, reported, high, disease, analysis, clinical, found, age, years, associated, showed, different, compared, risk, levels\\ 
    \hline
    \end{tabular}}
    \caption{top-10 words of topics learned by LDA on Pubmed dataset.}
    \label{tab:to}
    
\end{table}

In Table \ref{tab:rouge}, we further compare the mean score of ROUGE-1~\cite{lin2003automatic} F1 and ROUGE-2 F1 of the oracle summary, and summary based on the generated topics among all datasets used in our experiments.
\begin{table}[!hbt]
\scriptsize
    \centering
    \begin{tabular}{c|ccccc}
    \hline
    \bf{Dataset}& \bf{Oracle Summary}& \bf{Summary with Topics} \\
    \hline
    CNN/DM&0.811&0.174\\
    ArXiv&0.826&0.169\\
    PubMed-Long&0.845& 0.184\\
    PubMed-Short&0.847&0.187\\
    CORD-19 & 0.861&0.188\\
    S2ORC&0.841&0.193\\
    \hline
    \end{tabular}
     \caption{The mean score of ROUGE-1 F1 and ROUGE-2 F1 between different summaries with the gold summary, averaged on all documents.}
    \label{tab:rouge}
\end{table}
\begin{table}[!hbt]
    \centering
    \scriptsize
    \resizebox{0.4\textwidth}{30mm}{
    \begin{tabular}{|p{1.0\linewidth}|}
    
    \hline
     \textbf{Oracle summary:} this case report illustrates three learning points about cervical fractures in ankylosing spondylitis, and it highlights the need to manage these patients with the neck initially stabilised in flexion. We describe a case of cervical pseudoarthrosis that is a rare occurrence after fracture of the cervical spine with ankylosing spondylitis. This went undetected until the development of myelopathic symptoms many months later. The neck was initially stabilised in flexion using tongs, and then slowly extended before anterior and posterior fixation was performed. (\textbf{Mean score on ROUGE-1 F1 and ROUGE-2 F1: 0.994})\\
     \hline
     \textbf{Summary based on topic words:} 
     A patient's neurological condition may be made worse by \textcolor{red}{extension} of the \textcolor{red}{neck}, as the \textcolor{red}{spinal cord} may be compromised by the angle that is formed between the upper and lower \textcolor{red}{rigid} bony segments of the \textcolor{red}{cervical spine}.
     Over the previous 5 \textcolor{red}{weeks}, he had been experiencing increasing, although intermittent, symptoms including: sharp pains in the \textcolor{red}{posterior} aspect of his \textcolor{red}{neck} with \textcolor{red}{head movement}, abdominal pain and \textcolor{red}{paraesthesia} with \textcolor{red}{numbness} of his fingers and toes.
     Certainly significant trauma to a rigid and \textcolor{red}{osteoporotic spine} will cause \textcolor{red}{fracture}, and then the effect of instability at the \textcolor{red}{fracture} site (the fused \textcolor{red}{spinal} segments can be thought of as a long bone) will produce a pseudoarthrosis. (\textbf{Mean score on ROUGE-1 F1 and ROUGE-2 F1 : 0.172})\\
     \hline
     \textbf{Top-topic words:} cervical, spine, neck, ankylosing, fractures spondylitis, fixation, spinal, stabilised, anterior, posterior, flexion, fracture, c7, trauma, traction, paraesthesia, weakness, immobilisation, immobilised, bony, limb, head, cord, post-operatively.....\\ 
    \hline
    \end{tabular}}
    \caption{An example document. High-frequency topic words that appeared in sentences are marked with a red color.}
    \label{tab:example}
\end{table}
It shows a much lower rouge score on all datasets for summaries using generated topics, which indicates a semantic gap between the latent topics and the gold summary.
The latent topics would guide the method to select sentences that are topically similar to the posterior distribution of documents, rather than informative sentences, that cover the semantics in the gold summary.

\section{Method}
\begin{figure*}
  \centering
      \includegraphics[width=0.95\linewidth]{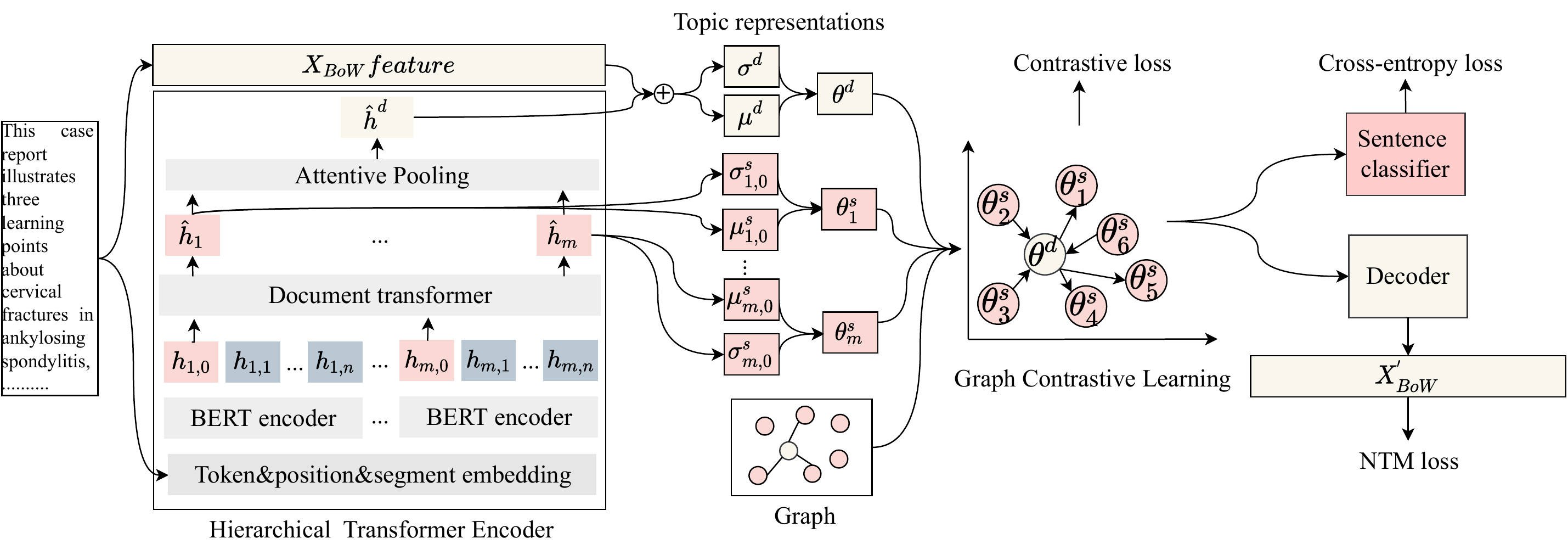}
       \caption{The model architecture of GRETEL}
\label{fig:model}
\end{figure*}
To address the aforementioned limitations of existing methods, we propose our method GRETEL, to better capture and incorporate the global semantics to improve PLMs, for long document extractive summarization.
Given $u$ sentences $\{s_1,\cdots,s_u\}$ of a document $i$ from the corpus $D$, extractive summarization aims to select $v$ informative sentences from $u$ sentences ($v \ll u$) as the summary $S$ for the document $i$. 
This task can be formulated as a binary sentence classification problem. 
We assign label $y_{i,j}=1$ to sentence $s_{i,j} (j \in \{1,\cdots,u\})$ for the summary, or $y_{i,j}=0$ otherwise.

As shown in Figure \ref{fig:model}, different from previous methods, we leverage the contextual representations from PLMs, and gold summary to guide the topic inference.
To this end, we first employ the hierarchical transformer encoder (HTE) to fully encode the global context of long documents, and then design the supervised graph contrastive loss, to push close the document topic distribution and topic distributions of salient sentences.
This helps our method to capture better global semantics, that effectively distinguish between salient and non-salient sentences, according to their contextual and semantic connections to the gold summary.

\subsection{Hierarchical Transformer Encoder}
To fully encode the document contents, especially for long documents, we propose to use a the Hierarchical Transformer Encoder (HTE) based on blocks with two modules: the block transformer encoder and the document transformer.

\textbf{Block transformer encoder}. We first split the document $d=\{blk_1,blk_2,\cdots,blk_m\}$ into $m$ blocks with fixed length, in which each block $blk_l=\{x_{l,0},\cdots,x_{l,n}\} (l \in \{1,\cdots,m\})$ has $n$ tokens.
Subsequently, each token $x_{l,p}(p \in \{1,\cdots,n\})$ at block $blk_l$ is represented by the vector $E_{l,p}$, which is the sum of the token embedding, the block embedding, and the position embedding.
Take $E_l$ as the input embedding, the contextual representations of tokens in each block $blk_l$ can be learned by the PLMs based transformer encoder: $h_{l}= BERT(E_l)$.
Following previous studies~\cite{liu2019text}, we insert the [CLS] and [SEP] tokens at the start and end of each sentence in the block.
We consider the representations of [CLS] tokens as the contextual representations of their corresponding sentences: $h^s=\{h^s_1,\cdots,h^s_u\}$, which capture the local contextual semantic in each block. 

\textbf{Document transformer encoder}. To further model the correlations among intra-block, we stack the document transformer encoder on $h^s$ to yield the document context-aware sentence representations: $\tilde{h}^s = Transformer(h^s)$.
To denote the position of each block, we add the block position embedding~\cite{vaswani2017attention} to $h^s$.
Finally, we use the pooling layer to generate the document representation $h^d$ based on $\tilde{h}^s$.

\subsection{Graph Contrastive Topic Model}
Next, we introduce the graph contrastive topic model, to capture global semantics empowered by the semantic information from HTE and the gold summary. 
It consists of a probabilistic topic encoder with HTE and supervised graph contrastive learning and a probabilistic decoder.

\subsubsection{Probabilistic Topic Encoder Enhanced with HTE}
We assume that $\theta^s$ and $\theta^d$ refer to sentence topic distributions and document topic distribution, $\beta$ represents the topics (topic word distributions in the vocabulary), and $X_i$ is the BoW feature of document $i$.
Different from existing methods~\cite{wang2020friendly, aralikatte2021focus, fu2020document} considering only the BoW features, we further employ the representations from HTE to leverage the semantic and syntactic dependencies among words to generate more coherent topics and topic distributions for documents and sentences.

For document topic distribution, we sample it from the logistic normal distribution \footnote{\scriptsize Following the previous method~\cite{srivastava2017autoencoding}, we use the logistic normal distribution to approximate the Dirichlet distribution.}. 
We first generate the mean and covariance of a multinomial distribution variable and then use the softmax activation function to convert it into the logistic normal distribution variable. 
Based on the contextual hidden representations from HTE and BoW features, for each document $i$, we have:
\begin{equation}
\small
\begin{split}
\tilde{h}^d_i &= f_{X}(X_i)+\tilde{h}^d_i\\
\mu^d_i &= f_{\mu}^d(\tilde{h}^d_i), \sigma^d = diag(f_{\sigma}^d(\tilde{h}^d_i ))\\
\theta^d_i &= softmax(\mu^d_i + {(\sigma^d_i)}^{\frac{1}{2}}\epsilon^d_i)\\
\label{eq:1}
\end{split}
\end{equation}
where $\epsilon^d_i \in N(0, I)$ is the sampled noise variable, ${f}_{\mu}^d$ and ${f}_{\sigma}^d$ are the feed-forward neural networks which takes input as the BoW feature $X_i$ and the contextual hidden representations $\tilde{h}^d_i$ of document $i$ from HTE respectively.

The sentence topic distribution is sampled with only the document context-aware hidden representations of sentences from HTE:
\begin{equation}
\small
\begin{split}
\mu^s_{i,j} &= f_{\mu}^s(\tilde{h}^s_{i,j}), \sigma^s_{i,j} = diag(f_{\sigma}^s(\tilde{h}^s_{i,j}))\\
\theta^s_{i,j} &= \mu^s_{i,j} + {(\sigma^s_{i,j})}^{\frac{1}{2}}\epsilon^s_{i,j}\\
\label{eq:2}
\end{split}
\end{equation}
where $\epsilon^s_{i,j} \in N(0, I)$ is the sampled noise variable, $f_{\mu}^s$ and $f_{\sigma}^s$ are the feed-forward neural networks which takes the same input as the representation from HTE for the sentence $j$ in the document $i$.
Notice that the BoW features for sentences are not considered since they would be too sparse to introduce external noise.

\subsubsection{Supervised Graph Contrastive Learning}
Although the posterior topic distributions can now utilize the sequential dependencies of words from HTE, they still cannot distinguish between important and redundant topics without the semantic information from the gold summary.
Thus, to fill the gap between the posterior topic distributions from NTMs and the key semantics in the gold summary, we propose the supervised graph contrastive learning to explicitly guide the document topic and sentence topic representations with the gold summary.

\textbf{Graph Construction}. For each document, we first build the graph $G=\{V, E\}$ with nodes $V$ as the document and all its sentences.
Edges $E$ can be represented by the adjacency matrix $A$, in which the edge between two nodes $(i, j)$ is defined as:
\begin{equation}
\small
   \begin{aligned}
A_{i,j} = \begin{cases} 1, & \textit{i is the document node}, \textit{ j is the}\\ 
&\textit{sentence node, and j} \in S^+\\
                         1, & i=j\\
                         0, & otherwise
          \end{cases}
\label{eq:a}
\end{aligned}
\end{equation}
where $S^+$ is the oracle summary of each document which has the maximum semantic similarity with the gold summary.
Notice that the graph is a bipartite graph with only connections between the document and sentences.

\textbf{Graph Contrastive Representation Learning}. Based on the bipartite graph embedded with the supervision information,
we argue that the representation of the document should be similar to representations of informative sentences in the oracle summary and dissimilar to representations of redundant sentences that are not mentioned.
Therefore, we design the following loss for the graph contrastive representation learning:
\begin{equation}
\small
\mathcal{L}_{con} = -\frac{1}{|V|}\sum_{i=1}^{|V|}log(\frac{\sum_{0<A_{i,j}}-A_{i,j}cos(x_i,x_j)}{\sum_{A_{i,j}=0}cos(x_i,x_j)})
\label{eq:g}
\end{equation}
where $|V|$ is the number of nodes, $cos$ denotes the cosine similarity, and $x_{i} \in {\theta^d_i, \theta^s_i}$ means the features of node $i$.
This loss explicitly pushes close the topic distributions of the nodes with connections in the graph, i.e., the document node and the sentence node in the oracle summary, and pulls away otherwise. 
It guides the model to learn more discriminative document and sentence distributions that are semantically related to the gold summary.

\subsection{Probabilistic Decoder}
Based on sampled sentences and document representations, we use the probabilistic decoder to generate the observed words and predict the labels of sentences in each document. 
For each document $i$, we assume the $v$-th word $w^d_{i,v}$ is generated from the multinomial distribution based on the dot product of the document representations and topics:
\begin{equation}
\small
p(w^d_{i,v}|(\theta^{d}_i,\beta);\Phi) = Mult([\theta^{d}_i\cdot\beta])
\label{eq:6}
\end{equation}
where $\Phi$ is the parameter set of the probabilistic decoder. The topics $\beta$ are randomly initialized. We assume the $j$-th sentence label $\tilde{y}_{i,j}$ of document $i$ is generated from the feed-forward neural network $f_y$ with the sigmoid activation function, based on the sentence representation $\theta^s_{i,j}$:
\begin{equation}
\small
p(\tilde{y}_{i,j}|\theta_{i,j}^s;\Phi) = f_y(\theta_{i,j}^s)
\label{eq:7}
\end{equation}
Since we fill the gap between the approximation and true global semantics with supervised contrastive learning based on both the contextual representations from HTE and BoW features,
our method allows us to directly use the sentence topic representations to predict the labels of sentences, without any further distillation or fusion.

\subsection{Optimization}
We optimize the loss function from both graph contrastive topic modeling and extractive summarization to support joint inference.
The final loss of GRETEL is the sum of the evidence lower bound (ELBO) and the graph contrastive loss:
\begin{equation}
\small
\mathcal{L} = \mathcal{L}(\Theta, \Phi; X_i) + \eta \mathcal{L}_{con}
\label{eq:10}
\end{equation}
where $\eta$ is the parameter to control the sensitivity of the contrastive normalization, $\mathcal{L}_{con}$ is the contrastive loss,
and $\mathcal{L}(\Theta, \Phi; X_i)$ is:
\begin{equation}
\small
\begin{split}
\mathcal{L}(\Theta, \Phi; X_i) &= \mathbb{E}_{q_{\Theta}(\theta_{i}^s|\tilde{h}^s_{i})}[log P_{\Phi}(y_i|\theta_{i}^s)]\\
&+\mathbb{E}_{q_{\Theta}(\theta^d_i|X_i,\tilde{h}^d_i)}[log P_{\Phi}(w_i|\theta^d_i,\beta)]\\
&-D_{KL}[q_{\Theta}(\theta^d_i|X_i,\tilde{h}^d_i)||P_{\Phi}(\theta^d_i)]
\label{eq:8}
\end{split}
\end{equation}
where $y_i$ is the ground truth labels of sentences in document $i$.
The ELBO is composed of three terms, including the sentence label prediction loss for the extractive summarization, the word reconstruction loss of the neural topic modeling,
and the KL divergence between the variational posterior and the prior of $\theta^d$, which uses the prior $P_{\Phi}(\theta^d|\alpha)$ to normalize the document topic representations. 

\section{Experimentation Details} 
In this section, we present the details of the datasets used, evaluation metrics, and different baselines.
\paragraph{Datasets.}
To evaluate the effectiveness of GRETEL, we conducted experiments on four benchmark datasets and two biomedical domain-specific long document datasets:
1) CNN/DM~\cite{hermann2015teaching}: a commonly used news dataset;
2) Arxiv~\cite{cohan2018discourse}: a dataset containing long scientific documents from the Arxiv website;
3) PubMed-Long~\cite{cohan2018discourse}: a  dataset containing long scientific documents from biomedicine;
4) PubMed-Short~\cite{zhong2020extractive}: adapted PubMed-Long to use only the introduction of the document as input and filter noisy documents;
5) CORD-19~\cite{wang2020cord}: an openly released dataset including long biomedical scientific papers related to COVID-19. We use the version of the dataset which was released on 2020-06-28~\cite{bishop2022gencomparesum,xie2022pre};
6) S2ORC~\cite{lo2020s2orc}: a publicly released dataset that includes long scientific papers from several domains. We sample a random subset of articles from only the biomedical domain~\cite{bishop2022gencomparesum,xie2022pre}.
We show the statistics of the datasets in Table \ref{tab:statistics}.
We use abstracts of documents as the gold summary.
For CNN/DM and Arxiv, we extract 3 and 7 sentences respectively to formulate the final summary, following previous methods~\cite{zhong2020extractive}.
For the remaining datasets, we extract 6 sentences to formulate the summary~\cite{bishop2022gencomparesum,xie2022pre}.
\begin{table}[!hbt]
\scriptsize
    \centering
    \begin{tabular}{c|ccccc}
    \hline
    \bf{Dataset}& \bf{Train}& \bf{Valid}& \bf{Test} & \bf{Avg len}&\bf{Ext} \\
    \hline
    CNN/DM&287,226&13,368&11,490&757&3\\
    ArXiv&203,037&6,436&6,440&5,038&7\\
    PubMed-Long&119,924& 6,633&	6,658&3,235&6\\
    PubMed-Short&83,233&4,676&5,025&444&6\\
    CORD-19 & 31,505 & 6,299&4,200&3,324&6\\
    S2ORC&47,782&9,556&	6,371&2,631&6\\
    \hline
    \end{tabular}
     \caption{Statistics of datasets. Ext denotes the number of sentences extracted in the final summary.}
    \label{tab:statistics}
\end{table}
\subsection{Implementation Details}
\label{imp}
Our method is implemented by Pytorch and Huggingface~\cite{wolf2020transformers}. 
We investigated the RoBERTa~\cite{liu2019roberta} implemented in Huggingface as the encoder. 
We use the base size of it.
We set the learning rate to 2e-3, dropout rate to 0.0, warmup steps to 5000, topic number between$\{100,200,300,400,500\}$, the parameter to control the negative samples of the contrastive loss $\gamma$ to 1, and the weight parameter $\eta$ to 0.5. 
We set the hidden size of the transformer in HTE to 768.
Due to the memory limitations of GPU, we set the max tokens of input documents as 6000.
We train the model with 50000 steps and save the model checkpoint at every 1000 steps. 
We select the best checkpoint according to the loss in the validation and report the results in the test. 
To extract the sentence label for training the model, we use the greedy search algorithm~\cite{nallapati2017summarunner} to select the oracle summary of each document, via maximizing the ROUGE-2 score against the gold summary.
we use the pyrouge\footnote{https://github.com/andersjo/pyrouge.git} to calculate the ROUGE~\cite{lin2004rouge} metric.

\paragraph{Baselines and Metrics.}
We compare our model with SOTA extractive summarization methods including: 
1) PLMs based methods: BERTSum~\cite{liu2019text} and MatchSum~\cite{zhong2020extractive}; 2) PLMs based models for long documents:  HIBERT~\cite{zhang2019hibert}, ETCSum~\cite{narayan2020stepwise}, SSN-DM~\cite{cui2021sliding}, GBT-EXTSUM~\cite{grail2021globalizing}, Longformer-Ext~\cite{beltagy2020longformer},
Reformer-Ext~\cite{kitaev2019reformer},
BERTSum+SW~\cite{liu2019text} which uses the BERTSum to sequentially encode the full context with the sliding window; 
3) topic enhanced transformer method: Topic-GraphSum~\cite{cui2021topic}, which is the only PLMs-based model with topic modeling for extractive summarization. 
Following ~\cite{liu2019text}, 
we report the unigram (ROUGE-1), bigram F1 (ROUGE-2), and the longest common subsequence (ROUGE-L) between the generated summary and the gold summary.
\begin{table*}[!hbt]
\scriptsize
    \centering
    \setlength\tabcolsep{3pt}
   \resizebox{\textwidth}{21mm}{
    \begin{tabular}{c|c|c|c|c|c|c|c|c|c|c|c|c|c|c|c|c|c|c}
    \hline
    \bf{Datasets}  & \multicolumn{3}{c|}{\bf{CNN/DM}} & \multicolumn{3}{c|}{\bf{Arxiv}}& \multicolumn{3}{c|}{\bf{PubMed-Long}}&\multicolumn{3}{c|}{\bf{PubMed-Short}} & \multicolumn{3}{c|}{\bf{CORD-19}} & \multicolumn{3}{|c}{\bf{S2ORC}}\\ 
    \hline
    Metrics&	R-1&	R-2&	R-L&	R-1&R-2&	R-L&	R-1&	R-2&	R-L&R-1&	R-2&	R-L&	R-1&	R-2&	R-L&R-1&	R-2&	R-L\\
    \hline
     LEAD&	40.11&17.54&36.32&33.66&8.94&22.19&36.19&11.82&32.96&37.58&12.22&	33.44&	32.40&8.97&29.30&36.62&16.57&33.11\\
    ORACLE&	56.22&33.74&52.19&53.88&23.05&44.90&50.26&28.32&46.33&45.12&20.33&	40.19&	46.20&22.86&42.08&	58.34&	34.48&	54.36\\
    \hline
    BERTSum&43.25&20.24&39.63&41.24&13.01&36.10&41.09&15.51&36.85&41.05&14.88&36.57&36.25&10.83&32.85&40.53&16.31&37.50\\
    MatchSum&44.41&20.86&40.55&-&-&-&-&-&-&41.21&14.91&	36.75&-&-&-&-&-&-\\
    \hline
    Topic-GraphSum&44.02&20.81&40.55&44.03&18.52&32.41&45.95&20.81&33.97&-&-&-&-&-&-&-&-&-\\
    \hline
    HiBERT&42.37&19.95&38.83&-&-&-&-&-&-&-&-&-&-&-&-&-&-&-\\
    ETCSum&43.84&20.80&39.77&-&-&-&-&-&-&-&-&-&-&-&-&-&-&-\\
    Longformer-Ext&43.00&20.20&39.30&45.24&16.88&40.03&43.75&17.37&39.71
&42.03&16.08&38.01&43.61&16.27&39.39&47.73&22.67&44.03\\
    Reformer-Ext&38.85&16.46&35.16&43.26&14.68&38.10&42.32&15.91&38.26&41.67&15.78&37.88&42.32&16.11&38.87&46.12&21.55&43.21\\
    HETFORMER&44.55&20.82&40.37&-&-&-&-&-&-&-&-&-&-&-&-&-&-&-\\
    SSN-DM&-&-&-&45.03&19.03&32.58&46.73&21.00&34.10&-&-&-&-&-&-&-&-&-\\
    BERTSum+SW&43.78&20.65&39.67&47.86&19.17&42.50&46.36&19.67&42.49&42.07&15.10&37.29&42.51&15.72&38.58&46.21&19.73&43.01\\
    GBT-EXTSUM&42.93&19.81&39.20&48.08&19.21&42.68&46.87&20.19&42.68&-&-&-&-&-&-&-&-&-\\
    \hline
    GRETEL&$\textbf{44.62}^\dagger$&$\textbf{20.96}^\dagger$&$\textbf{40.69}^\dagger$&$\textbf{48.17}^\dagger$&$\textbf{20.31}^\dagger$&$\textbf{42.84}^\dagger$&$\textbf{48.20}^\dagger$&$\textbf{21.20}^\dagger$&$\textbf{43.16}^\dagger$&$\textbf{42.53}^\dagger$&$\textbf{16.55}^\dagger$&$\textbf{38.61}^\dagger$&$\textbf{43.91}^\dagger$&$\textbf{16.54}^\dagger$&$\textbf{40.01}^\dagger$&
    $\textbf{48.24}^\dagger$&$\textbf{23.34}^\dagger$&$\textbf{44.55}^\dagger$\\
    \hline
    \end{tabular}}
    \caption{ROUGE F1 results of different models on CNN/DM, Arxiv, PubMed-Long, PubMed-Short, CORD-19, and S2ORC under 5 times running. $\dagger$ means outperform the existing model with best performance significantly ($p < 0.05$). Part results are from~\cite{grail2021globalizing, zhong2020extractive, cui2021sliding}.}
    \label{tab:results1}
\end{table*}
\section{Results and Analysis}
A series of experiments were conducted to demonstrate the efficacy of the proposed method.

\subsection{Main Results}
We first present the ROUGE F1 results of different models on all datasets in Table \ref{tab:results1}, which shows that our method GRETEL outperforms all existing baseline methods in all datasets.
It demonstrates the superiority of our method GRETEL to other methods, via capturing better global semantics with the guidance of the gold summary and the leverage of contextual information and word features simultaneously.
Our methods and Topic-GraphSum both present superior performance over methods without the topic information, such as BERTSum and MatchSum, indicating the importance of modeling the global semantic information with the approximation of latent topics.
When comparing with Topic-GraphSum incorporating latent topics, our method yields better performance on all datasets.
This proves the benefit of the supervision from the gold summary and the integration of contextual representations to exploit the better global semantics.
It is also proved from the improvement of our method when compared with methods that encode full document contents but ignore the topic information, such as Longformer-Ext, SSN-DM et al.

Moreover, our methods and other PLMs-based methods that address the truncation issue to encode full contents, such as Longformer-Ext, SSN-DM et al, achieve better performance on all long document datasets, when comparing methods with the input length limit, such as BERTSum, and MatchSum.
It shows that the content loss can inhibit their ability to model the contextual information in the document, which also limits the employment of the contextual representations from existing methods in the topic generation.
On the contrary, for CNN/DM and PubMed-Short whose documents are relatively short, the improvement is insignificant, since truncating these short documents would not miss much context information.
\begin{table}[!hbt]
\scriptsize
    \centering
    \resizebox{0.48\textwidth}{11mm}{
    \begin{tabular}{c|c|c|c|c|c|c}
    \hline
    \bf{Datasets}  & \multicolumn{3}{c|}{\bf{PubMed-Long}} & \multicolumn{3}{c}{\bf{Arxiv}}\\ 
    \hline
    Metrics&	R-1&	R-2&	R-L&R-1&R-2&R-L\\
    \hline
    GRETEL&$\textbf{48.20}^\dagger$&$\textbf{21.20}^\dagger$&$\textbf{43.16}^\dagger$&$\textbf{48.17}^\dagger$&$\textbf{20.31}^\dagger$&$\textbf{42.84}^\dagger$\\
    W/O HTE&45.97&20.13&40.22&45.44&16.53&40.15\\
    W/O Topic&47.61&20.89&42.86&47.48&19.97&42.65\\
    W/O Contras&47.65&20.96&42.92&47.95&20.02&42.76\\
    W/O Context&48.01&20.99&43.08&47.89&20.13&42.77\\
    W/O Document&47.61&21.02&43.09&48.11&20.18&42.80\\
    \hline
    \end{tabular}
    }
    \caption{ROUGE F1 results of our model under different settings on PubMed-Long and Arxiv.}
    \label{tab:abla}
\end{table}
\subsection{Ablation Study} 
We further verify the attribution of each component to the performance improvement of GRETEL in this section, as shown in Table \ref{tab:abla}.
It presents the results of our method including 1) W/O HTE replacing HTE with RoBERTa, 
2) W/O Topic removing the loss of neural topic modeling, 3) W/O Contras removing the graph contrastive loss, 3) W/O Context without considering the contextual representations from PLMs in generating document topic representations, and
4) W/O Document without the document transformer layer to propagate information between blocks.
We can observe that each component contributes to the performance of the model to a different degree.
Among all the components, HTE is the most important one for improvement, which shows the importance of encoder the full contents, when introducing contextual information into the topic modeling.
However, our method still outperforms Topic-GraphSum even without HTE, which demonstrates the superiority of the guidance from the gold summary during the topic generation in our method.

Furthermore, we show the impacts of different topic numbers on the performance of GRETEL in Table~\ref{tab:topic} on both the long document dataset PubMed-Long dataset and the short document dataset CORD-19.
It shows that the performance of our method generally increases with the growing number of topics on PubMed-Long, while it soon achieves the best of 300 topics on CORD-19 since it contains fewer topics with relatively shorter content and much fewer documents.
\begin{table}[!hbt]
\scriptsize
    \centering
    \begin{tabular}{c|c|c|c|c|c|c}
    \hline
    \bf{Datasets}  & \multicolumn{3}{c|}{\bf{PubMed-Long}} & \multicolumn{3}{c}{\bf{CORD-19}}\\ 
    \hline
    Metrics&	R-1&	R-2&	R-L&R-1&R-2&R-L\\
    \hline
    K=100&47.10&20.14&42.19&43.53&15.92&38.89\\
    K=200&47.53&20.38&42.42&43.80&16.25&39.41\\
    K=300&47.94&20.89&42.94&\textbf{43.91}&\textbf{16.54}&\textbf{40.01}\\
    K=400&48.15&21.12&43.24&43.85&16.42&39.94\\
K=500&$\textbf{48.20}$&$\textbf{21.20}$&$\textbf{43.26}$&43.86&16.45&40.01\\
    \hline
    \end{tabular}
    \caption{ROUGE F1 results of our model with a different number of topics on PubMed-Long and CORD-19.}
    \label{tab:topic}
\end{table}
\subsection{Topic Analysis}
To verify the quality of our generated topics, we further evaluate the NPMI~\cite{lau2014machine} score in Table~\ref{tab:npm} by our methods and the classical topic model LDA.
It clearly shows that our method can learn more coherent topics compared with LDA.
Moreover, our method without contextual information (W/O Contextual) and the supervision (W/O Contras) all outperform LDA and underperforms our method with both features.
It demonstrates that both the contextual information and the supervision from the gold summary are helpful to exploit meaningful and salient topics for a better approximation of global semantics.
\begin{table}[!hbt]
\scriptsize
    \centering
    
    \begin{tabular}{c|c|c|c|c}
    \hline
    \bf{Datasets} & \multicolumn{2}{c|}{\bf{CORD-19}} & \multicolumn{2}{c}{\bf{PubMed-Long}}\\ 
    \hline
    Model&K=100&K=200&K=100&K=200\\
    \hline
    LDA&0.18&0.16&0.14&0.19\\
    GRETEL&\bf{0.25}&\bf{0.21}&\bf{0.23}&\bf{0.26}\\
    W/O Contras&0.23&0.20&0.22&0.24\\
    W/O Contextual&0.20&0.18&0.18&0.20\\
    \hline
    \end{tabular}
     \caption{NPMI score of different models on CORD-19 and PubMed-Long using different numbers of topics.}
    \label{tab:npm}
\end{table}
\begin{table}[!hbt]
\scriptsize
    \centering
    \scriptsize
    \setlength\tabcolsep{3pt}
    \resizebox{0.49\textwidth}{80mm}{
    \begin{tabular}{|p{0.1\linewidth}|p{0.9\linewidth}|}
    \hline
 \textbf{Gold}& 
 A 53-year-old \textcolor{blue}{man} with steroid dependent rheumatoid arthritis presented with fever and serious articular drainage.
 Oral antibiotics were initially prescribed. Subsequent hemodynamic instability was attributed to septic \textcolor{blue}{shock}. Further evaluation revealed a pericardial effusion with tamponade.
  Pericardiocentesis of purulent fluid promptly corrected the hypotension. Proteus mirabilis was later \textcolor{blue}{isolated} from both the \textcolor{blue}{infected} \textcolor{blue}{joint} and the pericardial fluid.
  This is the first report of combined proteus mirabilis septic arthritis and purulent pericarditis.
  It documents the potential for atypical transmission of
  gram-negative pathogens, to the pericardium, in \textcolor{blue}{patients} with a high likelihood of preexisting pericardial \textcolor{blue}{disease}.
  In immunocompromised \textcolor{blue}{patients}, the typical signs and symptoms of pericarditis may be absent, and the \textcolor{blue}{clinical} presentation of pericardial tamponade may be misinterpreted as one of septic \textcolor{blue}{shock}. 
  This case underscores the value of a careful \textcolor{blue}{physical} \textcolor{blue}{examination} and proper interpretation of ancillary studies. It further illustrates the importance of initial antibiotic selection and the need for definitive \textcolor{blue}{treatment} of septic arthritis in immunocompromised \textcolor{blue}{patients}.\\
  \hline
\textbf{Our}& \begin{tcolorbox}[
 colback=red!50, 
 coltext=black, 
 sharp corners, 
 colframe=black, 
 boxrule=0pt, 
 boxsep=1pt,left=2pt,right=2pt,top=0pt,bottom=0pt,after skip=0pt,before skip=0pt
 ]ID 3: We report a case of purulent pericarditis with pericardial tamponade masquerading as septic \textcolor{blue}{shock} related to proteus mirabilis septic arthritis. \end{tcolorbox}
 \begin{tcolorbox}[
 colback=red!50, 
 coltext=black,
 sharp corners, 
 colframe=black,
 boxrule=0pt,
 boxsep=1pt,left=2pt,right=2pt,top=0pt,bottom=0pt,after skip=0pt,before skip=0pt
 ]ID 4: A 53-year-old \textcolor{blue}{man} with long-standing, steroid-dependent rheumatoid arthritis complained of a painful, swollen, left elbow with purulent drainage emanating from what appeared to be a small ulceration.\end{tcolorbox} \begin{tcolorbox}[
 colback=red!30, 
 coltext=black, 
 sharp corners, 
 colframe=black, 
 boxrule=0pt, 
 boxsep=2pt,left=2pt,right=2pt,top=0pt,bottom=0pt,after skip=0pt,before skip=0pt
 ]ID 0: Septic arthritis is a well recognized occurrence in \textcolor{blue}{patients} with steroid dependent rheumatoid arthritis .1 \textcolor{blue}{treatment} includes broad-spectrum antibiotics usually accompanied by surgical or needle drainage of the \textcolor{blue}{joint} .2 while pericardial effusions are common in \textcolor{blue}{patients} with rheumatologic disorders, the development of purulent pericarditis with pericardial tamponade is rare.\end{tcolorbox} \begin{tcolorbox}[
 colback=red!30, 
 coltext=black, 
 sharp corners, 
 colframe=black, 
 boxrule=0pt, 
 boxsep=2pt,left=2pt,right=2pt,top=0pt,bottom=0pt,after skip=0pt,before skip=0pt
 ]ID 75: This case also underscores the importance of appropriate antibiotic selection in the initial \textcolor{blue}{treatment} of immunocompromised \textcolor{blue}{patients} with \textcolor{blue}{infected} prosthetic \textcolor{blue}{joints}. \end{tcolorbox}\begin{tcolorbox}[
 colback=red!30, 
 coltext=black, 
 sharp corners, 
 colframe=black, 
 boxrule=0pt, 
 boxsep=2pt,left=2pt,right=2pt,top=0pt,bottom=0pt,after skip=0pt,before skip=0pt
 ]ID 30: While multiple \textcolor{blue}{blood} cultures were negative, articular and pericardial fluid cultures grew staphylococcus epidermidis and proteus mirabilis.\end{tcolorbox}\begin{tcolorbox}[
 colback=red!30, 
 coltext=black, 
 sharp corners, 
 colframe=black, 
 boxrule=0pt, 
 boxsep=2pt,left=2pt,right=2pt,top=0pt,bottom=0pt,after skip=0pt,before skip=0pt
 ] ID 68: In the setting of an \textcolor{blue}{infected} \textcolor{blue}{joint} prosthesis, fever, and immunosuppression, this \textcolor{blue}{patients} hemodynamic instability was initially ascribed to septic \textcolor{blue}{shock} and not to pericardial tamponade.\end{tcolorbox}\\
 \hline
\textbf{Baseline}& 
\begin{tcolorbox}[
 colback=red!50, 
 coltext=black, 
 sharp corners, 
 colframe=black, 
 boxrule=0pt, 
 boxsep=2pt,left=2pt,right=2pt,top=0pt,bottom=0pt,after skip=0pt,before skip=0pt
 ] ID 3: we report a case of purulent pericarditis with pericardial tamponade masquerading as septic shock related to proteus mirabilis septic arthritis.
 \end{tcolorbox}
 \begin{tcolorbox}[
 colback=red!50, 
 coltext=black, 
 sharp corners, 
 colframe=black, 
 boxrule=0pt, 
 boxsep=2pt,left=2pt,right=2pt,top=0pt,bottom=0pt,after skip=0pt,before skip=0pt
 ]
 ID 4: A 53-year-old man with long-standing, steroid-dependent rheumatoid arthritis complained of a painful, swollen, left elbow with purulent drainage emanating from what appeared to be a small ulceration. 
 \end{tcolorbox}
 \begin{tcolorbox}[
 colback=red!5, 
 coltext=black, 
 sharp corners, 
 colframe=black, 
 boxrule=0pt, 
 boxsep=2pt,left=2pt,right=2pt,top=0pt,bottom=0pt,after skip=0pt,before skip=0pt
 ]
 ID 12: On the first postoperative day, he was transferred to the medical service for the management of presumed septic shock. 
 \end{tcolorbox}
 \begin{tcolorbox}[
 colback=red!10, 
 coltext=black, 
 sharp corners, 
 colframe=black, 
 boxrule=0pt, 
 boxsep=2pt,left=2pt,right=2pt,top=0pt,bottom=0pt,after skip=0pt,before skip=0pt
 ]
 ID 25: A subxiphoid pericardiocentesis yielded 500 ml of purulent fluid with prompt normalization of the blood pressure. 
 \end{tcolorbox}
 \begin{tcolorbox}[
 colback=red!30, 
 coltext=black, 
 sharp corners, 
 colframe=black, 
 boxrule=0pt, 
 boxsep=2pt,left=2pt,right=2pt,top=0pt,bottom=0pt,after skip=0pt,before skip=0pt
 ]
 ID 0: While multiple blood cultures were negative, articular and pericardial fluid cultures grew staphylococcus epidermidis and proteus mirabilis.
 \end{tcolorbox}
 \begin{tcolorbox}[
 colback=red!5, 
 coltext=black, 
 sharp corners, 
 colframe=black, 
 boxrule=0pt, 
 boxsep=2pt,left=2pt,right=2pt,top=0pt,bottom=0pt,after skip=0pt,before skip=0pt
 ]
 ID 11: The early postoperative course, however, was remarkable for persistent fever, hypotension, and tachycardia.
 \end{tcolorbox}\\
\hline
\textbf{Topics}& T267: \textcolor{blue}{infected} congenital \textcolor{blue}{patients} loss \textcolor{blue}{blood} \textcolor{blue}{disorders} compared \textcolor{blue}{high} fig phase 
T433: \textcolor{blue}{treatment} \textcolor{blue}{infected} severity \textcolor{blue}{year} \textcolor{blue}{patients} crp tract area arm \textcolor{blue}{isolated} 
T446: \textcolor{blue}{joints} dna \textcolor{blue}{physical} risk observed tube lower \textcolor{blue}{examination} intravenous features
T153: \textcolor{blue}{disease} type \textcolor{blue}{exercise} vegf nerve deaths \textcolor{blue}{shock} \textcolor{blue}{joints} drugs lower
T108: type \textcolor{blue}{treatment} \textcolor{blue}{male} sd well use statistically specific post mice\\
\hline
    \end{tabular}}
    \caption{Example of extractive summarization conducted by our method on the PubMed-Long dataset. The gold summary is the abstract of the document. Sentences with deep color have a higher ROUGE score. Topic words are marked with the blue color.}
    \label{tab:case}
\end{table}
\begin{table}[!hbt]
    \centering
    
    \resizebox{0.49\textwidth}{20mm}{
    \begin{tabular}{llr}
    \toprule
    Sentence ID & Top-6 words\\
    \midrule
   \multirow{2}{*}{ID 3} & adequate mice \textcolor{blue}{patients} label understanding body \textcolor{blue}{infected} \textcolor{blue}{report} oxygen formation\\
   &\textcolor{blue}{disease} type \textcolor{blue}{exercise} vegf nerve deaths \textcolor{blue}{shock} \textcolor{blue}{joints} drugs lower\\
    \midrule
     \multirow{2}{*}{ID 4} & family \textcolor{blue}{report} presented followed obesity side j macrophages \textcolor{blue}{high} necrosis\\
     & \textcolor{blue}{treatment} \textcolor{blue}{infected} severity \textcolor{blue}{year} \textcolor{blue}{patients} crp tract area arm \textcolor{blue}{isolated}\\
    \midrule
     \multirow{2}{*}{ID 0} & \textcolor{blue}{treatment} \textcolor{blue}{infected} severity \textcolor{blue}{year} \textcolor{blue}{patients} crp tract area arm \textcolor{blue}{isolated}\\
     &type \textcolor{blue}{treatment} \textcolor{blue}{male} sd well use statistically specific post mice\\
     \midrule
     \multirow{2}{*}{ID 75} & \textcolor{blue}{treatment} \textcolor{blue}{infected} severity \textcolor{blue}{year} \textcolor{blue}{patients} crp tract area arm \textcolor{blue}{isolated}\\
     &\textcolor{blue}{treatment} adjacent medication different motor min height stroke like rate\\
     \midrule
     \multirow{2}{*}{ID 30} & et treatment lower diagnosis control observed could 7 number association\\
     &\textcolor{blue}{infected} congenital \textcolor{blue}{patients} loss \textcolor{blue}{blood} \textcolor{blue}{disorders} compared \textcolor{blue}{high} fig phase \\
     \midrule
     \multirow{2}{*}{ID 68} & type \textcolor{blue}{treatment} \textcolor{blue}{male} sd well use statistically specific post mice\\
     &effects performed revealed compared clinical observed diagnosis \textcolor{blue}{isolated} cm family\\
    \bottomrule
    \end{tabular}}
    \caption{Top 10 words of top 2 topics in sentences, which are selected into the summary.}
    \label{tab:st}
  \end{table}
\begin{figure}
  \centering
      \includegraphics[width=0.65\linewidth]{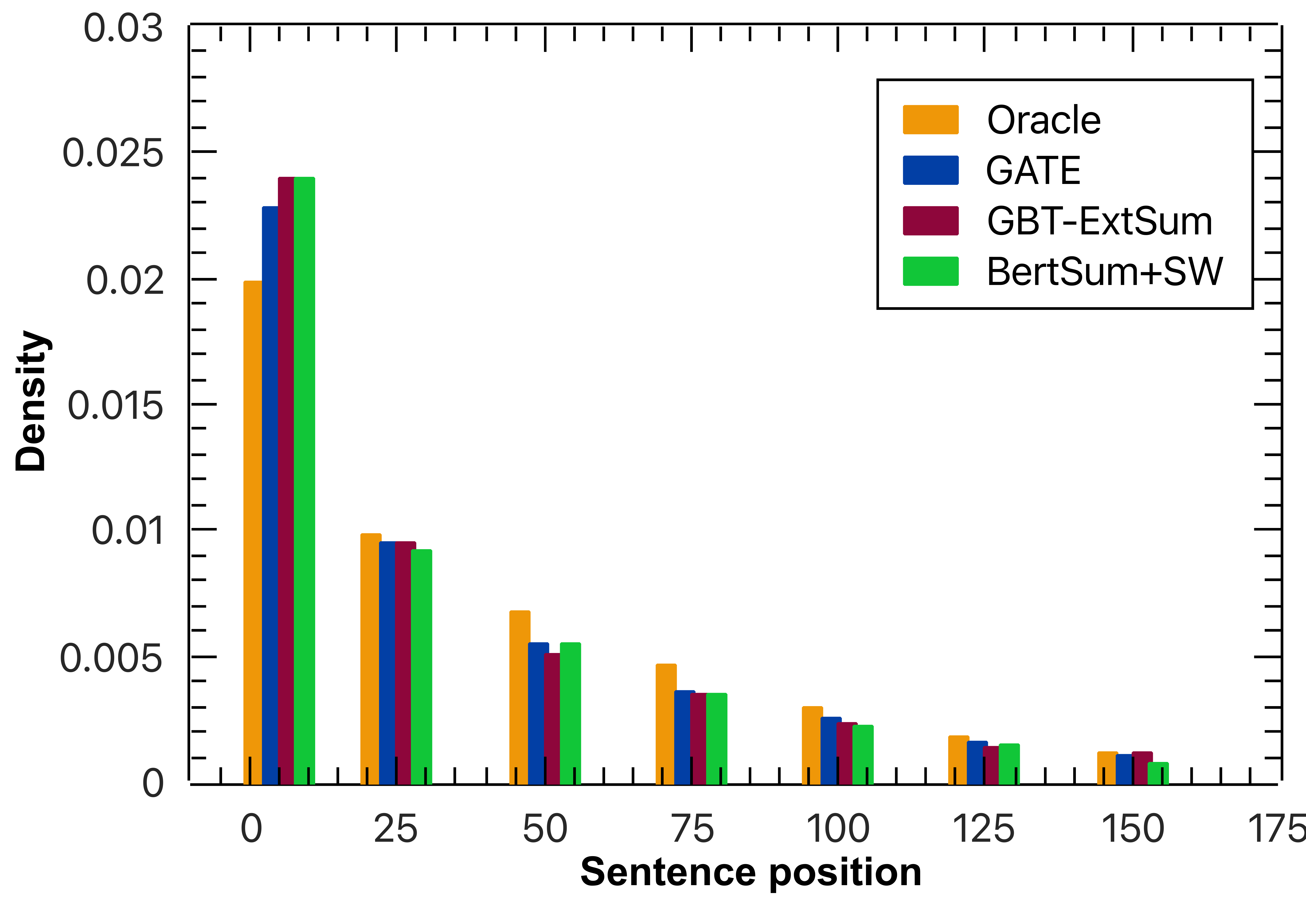}
       \caption{The position distribution of extracted sentences by different models on the PubMed-Long test set.}
\label{fig:position}
\end{figure}
\subsection{Case Study}
In Table \ref{tab:case}, we present the summaries of an example document generated by GRETEL and BERTSum, together with the top-5 topics (with the highest coherence) of the document from the PubMed-Long dataset.
In table \ref{tab:st}, we show the top-2 topics of selected sentences by our method for inclusion in the summary.
It shows that our method generates a more coherent summary that contains more salient sentences than the summary generated by BERTSum, due to the integration of a better approximation of global semantics in our method.
This is also proved by the selected sentences of our method are topically related to the captured topics about "treatment", "joints" and "infected", which are semantically similar to the meaning of the gold summary.

Moreover, the positions of our selected sentence vary in every part of the document while the sentences of BERTSum are all located in the former part of the document.
This is because the employment of HTE allows our method to encode the full contents of the document without truncation.
In Figure \ref{fig:position}, we further compare the position distribution of selected sentences by different models and the oracle summary on PubMed-Long.
The distribution of our method is the most similar to the oracle summary, which pays more attention to the latter sentences compared with other models.
\section{Conclusion}
In this paper, we propose a novel framework GRETEL for extractive summarization of long texts, that furnishes PLMs with the neural topic inference, to fully incorporate the local and global semantics.
Experimental results on both general and biomedical datasets show that our model outperforms existing state-of-the-art methods, and global semantics empowered by graph contrastive learning and PLMs can yield more discriminative sentence representations to select salient sentences, that are topically similar to the gold summary. 
For future work, we would explore the feasibility of extending this framework to abstractive and multi-document summarization tasks.
\section*{Acknowledgments}
This research is supported by the Alan Turing Institute and the Biotechnology and Biological Sciences Research Council (BBSRC), BB/P025684/1. 
We would like to thank Pan Du, Jennifer Bishop, and Guanghao Yang for their help and constructive comments.
\bibliography{anthology}
\bibliographystyle{acl_natbib}
\end{document}